\documentclass[a4paper]{article}%
\usepackage{hyperref}
\usepackage{amsmath}
\usepackage{graphicx}
\usepackage{amsfonts}
\usepackage{amssymb}
\usepackage{algorithm}
\usepackage{algorithmic}%

\begin{document}

\title{Estimation of Personalized Heterogeneous Treatment Effects Using Concatenation
and Augmentation of Feature Vectors}
\author{Lev V. Utkin, Mikhail V. Kots, Viacheslav S. Chukanov\\Peter the Great St.Petersburg Polytechnic University (SPbPU)\\St.Petersburg, Russia\\e-mail: lev.utkin@gmail.com, mkots96@gmail.com, kauter1989@gmail.com}
\date{}
\maketitle

\begin{abstract}
A new meta-algorithm for estimating the conditional average treatment effects
is proposed in the paper. The main idea underlying the algorithm is to
consider a new dataset consisting of feature vectors produced by means of
concatenation of examples from control and treatment groups, which are close
to each other. Outcomes of new data are defined as the difference between
outcomes of the corresponding examples comprising new feature vectors. The
second idea is based on the assumption that the number of controls is rather
large and the control outcome function is precisely determined. This
assumption allows us to augment treatments by generating feature vectors which
are closed to available treatments. The outcome regression function
constructed on the augmented set of concatenated feature vectors can be viewed
as an estimator of the conditional average treatment effects. A simple
modification of the Co-learner based on the random subspace method or the
feature bagging is also proposed. Various numerical simulation experiments
illustrate the proposed algorithm and show its outperformance in comparison
with the well-known T-learner and X-learner for several types of the control
and treatment outcome functions.


\end{abstract}

\section{Introduction}

One the most important problems in medicine is to choose the most appropriate
treatment for a certain patient which may differ from other patients in
her/his clinical or other characteristics \cite{Lu-Sadiq-etal-2017}. With the
increase of the amount of data and with the developing the electronic health
record concept in medicine, there is a growing interest to apply machine
learning methods to solve the problem of the most appropriate treatment by
estimating treatment effects directly from observational data. The main
peculiarity of observational data is that it contains past actions, their
outcomes, but without direct access to the mechanism which gave rise to the
action. Shalit at al. \cite{Shalit-etal-2017} give a clear example of
observational data, when we have patient characteristics, medications
(action), and outcomes, but we do not have complete knowledge of why a
specific action was applied to a patient.

By having observational data, average treatment effect (ATE) can be estimated,
which is the mean difference between outcomes of patients from the treatment
and control groups. However, one of the main peculiarities of many medical
application problems is that patients differ not only in background
characteristics, but also in how they respond to a particular treatment. It is
interesting to know what is the specific treatment effect for a given patient
characterizing, for example, gender, age, CT scan images. This leads to
heterogeneity of the treatment effect which is estimated in this case by
conditional average treatment effects (CATE). CATE is defined as the ATE
conditional on an patient feature vector
\cite{Green-Kern-2012,Hill-2011,Kallus-2016,Wager-Athey-2015}. A similar term
is the heterogeneous treatment effect (HTE). CATE are treatment effects at the
individual level as opposed to the ATE at the population level
\cite{Fan-Lv-Wang-2018}. The problem of computing CATE is very important in
personalized medicine \cite{Powers-etal-2017}. One of the challenging problems
in estimating CATE is that, in many real-world applications, we observe that
the number of control patients is much larger than the number of treated ones.
Another challenging problem is that, for every patient, we either observe its
outcome under treatment or control, but never both \cite{Kunzel-etal-2018a}. A
challenging problem is that causal inference is also challenging because data
from health care databases are typically noisy, high dimensional, and most
importantly, observational \cite{Wendling-etal-2018}.

Successful applications of machine learning to solving the problem of
computing CATE or HTE started a lot of corresponding models last years. The
well-known Lasso model was adapted to solving the CATE problem
\cite{Jeng-Lu-Peng-2018}. The CATE problem was represented as the SVM model
\cite{Zhao-etal-2012,Zhou-Mayer-Hamblett-etal-2017}. Athey and Imbens
\cite{Athey-Imbens-2016} provided a \textquotedblleft honest\textquotedblright%
\ model for estimation of the HTE. The main idea underlying the model is
splitting the training sample into two parts, one sample is used to construct
the partition of the data into subpopulations that differ in the magnitude of
their treatment effect, and another sample is used to estimate treatment
effects for each subpopulation. Deng et al. \cite{Deng-etal-2016} considered
two aspects in HTE understanding, one is to predict the effect conditioned on
a given set of side information or a given individual, the other is to
interpret the HTE structure and summarize it in a memorable way. The former
aspect can be treated as a regression problem, and the latter aspect focuses
on concise summarization and interpretation. Athey et al.
\cite{Athey-Tibshirani-Wager-2016,Athey-Tibshirani-Wager-2018} develop a
unified framework for the design of fast tree-growing procedures for tasks
that can be characterized by heterogeneous estimating equations. Zhang et al.
\cite{Zhang-Le-etal-2017} proposed a modification of the survival causal tree
method for estimating the HTE based on censored observational data. Xie et al.
\cite{Xie-Chen-Shi-2018} established the HTE detection problem as a false
positive rate control problem, and discussed in details the importance of this
approach for solving large-scale HTE detection problems. Alaa and Schaar
\cite{Alaa-Schaar-2018} considered guidelines by characterizing the
fundamental limits of estimating HTE, and establishing conditions under which
these limits can be achieved. The guiding principles for designing practical
HTE estimation algorithms were provided in the context of Bayesian
nonparametric inference. Bayesian additive regression trees, causal forest,
causal boosting, and causal MARS were numerically compared under condition of
binary outcomes by using the simulation experiments in
\cite{Wendling-etal-2018}. An orthogonal random forest as an algorithm that
combines orthogonalization with generalized random forests for solving the HTE
estimation problem was proposed in \cite{Oprescu-Syrgkanis-Wu-2018}. An
explanation of the HTE estimation based on the regression is considered in
\cite{Rhodes-2010}. Grimmer et al. \cite{Grimmer-etal-017} illustrated how the
weighting of ensemble-based methods can contribute to accurate estimation of
the HTE. McFowland III et al. \cite{McFowland-etal-2018} considered the
problem of the HTE estimation in the framework of anomaly detection. Kunzel et
al. \cite{Kunzel-etal-2018a} proposed new algorithms for estimating CATE using
transfer learning and neural networks. By taking into account the fact that
different CATE estimators provide quite different estimates, Kunzel et al.
\cite{Kunzel-etal-2018b} considered the application of a set of CATE
estimators. A lot of interesting approaches for estimating HTE can also be
found in
\cite{Chen-Liu-2018,Kallus-Puli-Shalit-2018,Kallus-Zhou-2018,Knaus-etal-2018,Levy-2018,Powers-etal-2017,Xie-Brand-Jann-2012}%
. This is only a small part of methods, algorithms and models proposed in last
years for estimating CATE.

A set of meta-algorithms or meta-learners for estimating CATE were considered
by Kunzel et al. \cite{Kunzel-etal-2018}. The algorithms consist of the base
learners (random forest, linear regression, etc.) dealing with control and
treatment groups separately and a meta-level which can be viewed as a function
of the base learners. We can pointed out the following meta-learners: the
T-learner \cite{Kunzel-etal-2018}, the S-learner \cite{Kunzel-etal-2018}, the
O-learner \cite{Wang-etal-2016}, the X-learner \cite{Kunzel-etal-2018}. There
are also the Q-learner and the A-learner
\cite{Laber-etal-2014,Murphy-etal-2007,Qian-Murphy-2011,Schulte-etal-2014},
which can be regarded as an approximate dynamic programming algorithm for
estimating optimal sequential decision rules from data or the well-known
dynamic treatment regimes, but not as meta-learners in the sense of the
two-step algorithms analyzed by Kunzel et al. \cite{Kunzel-etal-2018}.

One of the most efficient and interesting meta-algorithms is the X-learner
which consists of two main steps. The initial regression models are
constructed for the control and treatment groups separately at the first step.
New regression models of residuals between the obtained initial regression
models and data are constructed at the second step. These new regression
models are combined to get an estimate of CATE. Kunzel et al.
\cite{Kunzel-etal-2018} pointed out two main advantages of the X-learners.
First, it adapts to structural properties such as sparsity or smoothness of
the CATE. Second, it is effective when the number of patients in the control
group is much larger than in the treatment group.

We propose another meta-algorithm for estimating CATE which is based on the
concatenation of feature vectors from control and treatment groups and the
augmentation of the concatenated vectors. The concatenation operation is a
reason for calling the method as Co-learner. We assume that the number of
control patients is rather large, and a proper regression model can be
constructed for these patients. The first main idea underlying the
meta-algorithm is that we construct a new dataset consisting of concatenated
pairs of the control and treatment feature vectors such that the outcome for
every pair is determined as the difference of outcomes of the corresponding
feature vectors. The second idea is to extend the obtained concatenated
dataset by generating random control feature vectors for every treatment
vector in accordance with a special rule. Moreover, every new concatenated
synthetic feature vector is assigned by a weight which depends on the distance
between the treatment vector and the random control vectors. We assume that
the number of controls significantly exceeds the number of patients in
treatment group. Moreover, we propose a simple modification of the Co-learner,
which is on applying the well-known random subspace method \cite{Ho-1998} or
the feature bagging. The feature bagging is one of the ensemble-based methods
\cite{ZH-Zhou-2012}, which trains a set of individual models, for example,
Co-learners, on random samples of features instead of the entire feature set.

Various numerical experiments illustrate the outperformance of the Co-learner
as well as the Co-learner with the feature bagging in comparison with the
T-learner and the X-learner for some control and treatment output function. We
also consider cases when the proposed meta-models provide results which are
inferior to the T-learner and the X-learner.

The paper is organized as follows. A formal statement of the CATE estimation
problem is given in Section 2. The well-known meta-learner, including, the
T-learner, S-learner, and the X-learner are described in Section 3. Section 4
provides main ideas underlying the proposed Co-learner. An algorithm
implementing the Co-learner is given in the same section. Numerical
experiments illustrating the Co-learner and its comparison with the T-learner
and the X-learner are discussed in Section 5. The modification of the
Co-learner based the feature bagging with the corresponding numerical
experiments are given in Section 6. Concluding remarks are provided in Section 7.

\section{A formal problem statement}

Formally, let $\mathcal{D}=\{(X_{1}^{(T_{1})},y_{1}^{(T_{1})},T_{1}%
),...,(X_{n}^{(T_{n})},y_{n}^{(T_{n})},T_{n})\}$ be a training set where
$X_{i}^{(T_{i})}=(x_{i1}^{(T_{i})},...,x_{im}^{(T_{i})})\in\mathbb{R}^{m}$ is
the $m$-dimensional covariate or feature vector for a patient $i$; $T_{i}%
\in\{0,1\}$ is the treatment assignment indicator such that $T_{i}=0$
corresponds to the control group, and $T_{i}=1$ corresponds to the treatment
group; $y_{i}^{(T_{i})}\in\mathbb{R}$ is the observed outcome, for example,
time to death of the $i$-th patient or the blood sugar level, such that
$y_{i}^{(0)}$ and $y_{i}^{(1)}$ represent outcomes if $T_{i}=0$ and $T_{i}=1$,
respectively. It is assumed that the training set is a realization of $n$
independent random variables from a distribution $\mathcal{P}$. Denote subsets
of $\mathcal{D}$ corresponding to patients from the control and treatment
groups as $\mathcal{C}$ and $\mathcal{T}$, respectively, and $\mathcal{D}%
=\mathcal{C}\cup\mathcal{T}$. Suppose that numbers of feature vectors in
$\mathcal{C}$ and $\mathcal{T}$ are $n_{0}$ and $n_{1}$, respectively.

The causal effect of the treatment on a new patient $i$ with the feature
vector $X_{i}$, which allows us to make decision about the usefulness or
efficiency of the treatment can be estimated by the Individual Treatment
Effect (ITE) defined as $y_{i}^{(1)}-y_{i}^{(0)}$. Since the ITE cannot be
observed, then the causal effect is estimated as using the conditional average
treatment effect (CATE) defined as the expected difference between the two
potential outcomes as follows \cite{Rubin-2005}:
\[
\tau(X)=\mathbb{E}\left[  y_{i}^{(1)}-y_{i}^{(0)}|X_{i}=X\right]  .
\]

It is shown in \cite{Kunzel-etal-2018} that the best estimator for the CATE is
also the best estimator for the ITE.

The fundamental problem of computing the causal effect is that for each
patient in the training dataset, we only observe one of potential outcomes
$y_{i}^{(1)}$or $y_{i}^{(0)}$, but never both. A widely used assumption to
overcome the above problem, which is accepted in most models of estimating
CATE, is the assumption of unconfoundedness \cite{Rosenbaum-Rubin-1983}, i.e.,
the treatment assignment $T_{i}$ is independent of the potential outcomes for
$y_{i}$ conditional on $X_{i}$, which can be written as
\[
T_{i}\perp\{y_{i}^{(0)},y_{i}^{(1)}\}|X_{i}.
\]

An excellent explanation of the unconfoundedness assumption is provided by
Imbens \cite{Imbens-2004}. The assumption allows the untreated units to be
used to construct an unbiased counterfactual for the treatment group.

Another assumption regards the joint distribution of treatments and
covariates. It is called the overlap assumption and can be written as%
\[
0<\Pr\{T_{i}=1|X_{i}\}<1.
\]

It means that for each value of $X$, there is a positive probability of being
both treated and untreated. In other words, there is sufficient overlap in the
characteristics of treated and untreated patients to find adequate matches.

Taking into account the above assumptions, the CATE\ can be rewritten as
\[
\tau(X)=\mathbb{E}\left[  y_{i}^{(1)}|X_{i}=X\right]  -\mathbb{E}\left[
y_{i}^{(0)}|X_{i}=X\right]  .
\]

As pointed out by Wager and Athey \cite{Wager-Athey-2017}, the motivation
behind unconfoundedness is that nearby observations in the feature space can
be treated as having come from a randomized experiment, i.e., nearest-neighbor
matching and other local methods may be consistent for $\tau(X)$. This is a
very important property which will be used later.

Suppose that the outcomes can be expressed through the functions $g_{0}$ and
$g_{1}$ as follows:
\[
y^{(0)}=g_{0}(X)+\varepsilon,\ X\in\mathcal{C},
\]%
\[
y^{(1)}=g_{1}(X)+\varepsilon,\ X\in\mathcal{T}.
\]

Here $\varepsilon$ is a is a Gaussian noise variable such that $\mathbb{E}%
(\varepsilon)=0$. Hence, there holds
\begin{equation}
\tau(X)=g_{1}(X)-g_{0}(X). \label{HTE_concat_20}%
\end{equation}

\section{Meta-learners}

Kunzel et al. \cite{Kunzel-etal-2018} considered meta-algorithms which are
based on the so-called base learners, for example, random forests, neural
networks, etc. Meta-algorithms for estimating the HTE takes two steps. First,
it uses base learners to estimate the conditional expectations of the outcomes
given predictors under control and treatment separately. Second, it takes the
difference between these estimates.

\subsection{T-learner}

The T-learner \cite{Kunzel-etal-2018} is a simple procedure based on
estimating the control $g_{0}$ and treatment $g_{1}$ outcome functions by
applying a regression algorithm as:
\[
g_{0}(X)=\mathbb{E}\left[  Y^{(0)}|X_{i}=X\right]  ,\ g_{1}(x)=\mathbb{E}%
\left[  Y^{(1)}|X_{i}=X\right]  .
\]

\begin{figure}
[ptb]
\begin{center}
\includegraphics[
%%natheight=2.208700in,
%%natwidth=3.525800in,
height=1.8983in,
width=3.0173in
]%
{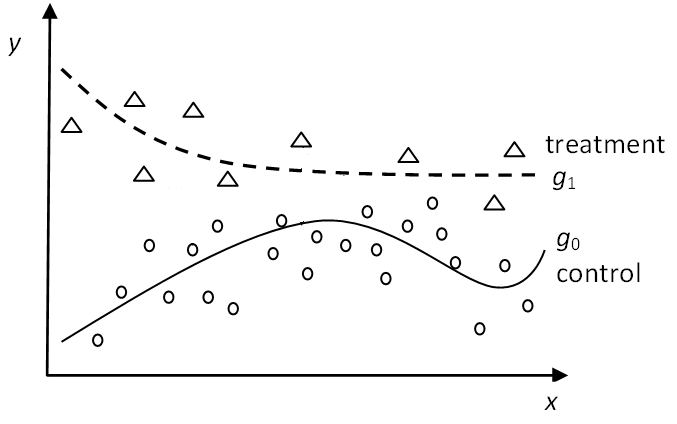}%
\caption{Functions $g_{0}$ and $g_{1}$ in accordance with the T-learner
algorithm}%
\label{fig:t_learner}%
\end{center}
\end{figure}

The CATE in this case is defined as the difference (\ref{HTE_concat_20}). An
example of the functions $g_{0}$ and $g_{1}$ is shown in Fig.
\ref{fig:t_learner}. The observed outcomes for patients from the control group
and for patients from the treatment group are represented by means of circles
and triangles, respectively. The solid and dashed curves are the functions
$g_{0}$ and $g_{1}$ corresponding to the control and treatment regressions, respectively.%

%TCIMACRO{\FRAME{ftbpFU}{3.0173in}{1.8983in}{0pt}{\Qcb{Functions $g_{0}$ and
%$g_{1}$ in accordance with the T-learner algorithm}}{\Qlb{fig:t_learner}%
%}{t_learner.png}{\special{ language "Scientific Word";  type "GRAPHIC";
%width 3.0173in;  height 1.8983in;  depth 0pt;  original-width 3.5258in;
%original-height 2.2087in;  cropleft "0";  croptop "1";  cropright "1";
%cropbottom "0";  filename 'T_learner.png';file-properties "XNPEU";}}}%
%BeginExpansion
%EndExpansion

\subsection{S-learner}

The S-learner was proposed by Kunzel et al. \cite{Kunzel-etal-2018} in order
to avoid some disadvantages of the T-learner. According to the algorithm of
the S-learner, the treatment assignment indicator $T_{i}$ is included as an
additional feature to the feature vector $X$, i.e., the training set is
modified as $\mathcal{D}^{\ast}=\{(X_{1}^{\ast},y_{1}),...,(X_{n}^{\ast}%
,y_{n})\}$, where $X_{i}^{\ast}=(X_{i},T_{i})\in\mathbb{R}^{m+1}$. Then the
outcome function $g(X,T)$ is estimated by using the training set
$\mathcal{D}^{\ast}$. The CATE is determined in this case as
\[
\tau(X)=g(X,1)-g(X,0).
\]

\subsection{X-learner}

\begin{figure}
[ptb]
\begin{center}
\includegraphics[
%%natheight=2.208700in,
%%natwidth=3.525800in,
height=1.9527in,
width=3.1038in
]%
{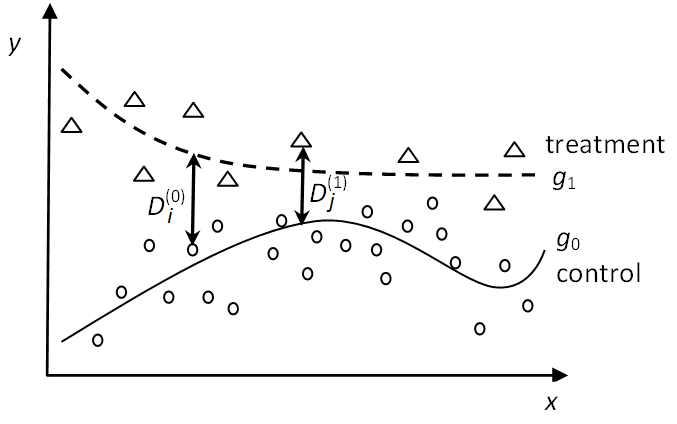}%
\caption{Imputed treatment effects in accordance with the X-learner algorithm}%
\label{fig:x_learner}%
\end{center}
\end{figure}

The X-learner \cite{Kunzel-etal-2018} is represented in the following three steps:

\begin{enumerate}
\item The outcome functions $g_{0}$ and $g_{1}$ are estimated by applying a
regression algorithm as:
\[
g_{0}(X)=\mathbb{E}\left[  y^{(0)}|X\right]  ,\ g_{1}(X)=\mathbb{E}\left[
y^{(1)}|X\right]  .
\]

\item Imputed treatment effects are computed as follows:
\[
D_{i}^{(1)}=y_{i}^{(1)}-g_{0}(X_{i}^{(1)}),\ \ D_{i}^{(0)}=g_{1}(X_{i}%
^{(0)})-y_{i}^{(0)}.
\]
The imputed treatment effects are shown in Fig. \ref{fig:x_learner} where the
observed outcomes for patients from the control group (circles) and for
patients from the treatment group (triangles) are depicted. The solid and
dashed curves are the functions $g_{0}$ and $g_{1}$ corresponding to the
control and treatment regressions, respectively. It can be seen from Fig.
\ref{fig:x_learner} that $D_{j}^{(1)}$ is the distance between a triangle (a
patient from the treatment group) and the solid curve, $D_{i}^{(0)}$ is the
distance between a circle (a patient from the control group) and the dashed curve.

\item Two regression functions $\tau_{1}(X)$ and $\tau_{0}(X)$ are estimated
for imputed treatment effects $D_{j}^{(1)}$ and $D_{i}^{(0)}$, respectively,
i.e., there hold
\[
\tau_{1}(X)=\mathbb{E}\left[  D^{(1)}|X\right]  ,\ \ \tau_{0}(X)=\mathbb{E}%
\left[  D^{(0)}|X\right]  .
\]
The estimation is carried out by a regression algorithm. The CATE\ is defined
as a weighted linear combination of the functions $\tau_{1}(X)$ and $\tau
_{0}(X)$ as
\[
\tau(X)=\alpha(X)\tau_{0}(X)+(1-\alpha(X))\tau_{1}(X).
\]
Here $\alpha(X)\in\lbrack0,1]$ is a weighting function which aims to minimize
the variance of $\tau(X)$. Kunzel et al. \cite{Kunzel-etal-2018a} propose to
choose the function to be constant and equal to the ration of treated patients.
\end{enumerate}

%

%TCIMACRO{\FRAME{ftbpFU}{3.1038in}{1.9527in}{0pt}{\Qcb{Imputed treatment
%effects in accordance with the X-learner algorithm}}{\Qlb{fig:x_learner}%
%}{x_learner.png}{\special{ language "Scientific Word";  type "GRAPHIC";
%width 3.1038in;  height 1.9527in;  depth 0pt;  original-width 3.5258in;
%original-height 2.2087in;  cropleft "0";  croptop "1";  cropright "1";
%cropbottom "0";  filename 'X_learner.png';file-properties "XNPEU";}}}%
%BeginExpansion
%EndExpansion

\section{The proposed Co-learner}

The main idea of the proposed meta-algorithm for estimating the CATE is to
consider another feature space which consists of the concatenated vectors
$Z_{ij}=X_{i}^{(1)}||X_{j}^{(0)}\in\mathbb{R}^{2m}$ such that $X_{j}^{(0)}%
\in\mathcal{C}$ and $X_{i}^{(1)}\in\mathcal{T}$. We denote the operation of
concatenation by means of symbol $||$. In other words, we produce feature
vectors consisting of two parts. The first part is a feature vector from the
treatment group, and the second part is a feature vector from the control
group. If outcomes $y_{i}^{(1)}$ and $y_{i}^{(0)}$ correspond to feature
vectors $X_{i}^{(1)}$ and $X_{j}^{(0)}$, respectively, then we will
characterize the vector $Z_{ij}$ by the difference $y_{i}^{(1)}-y_{i}^{(0)}$.
As a result, we get a new training set
\[
\mathcal{G}=\left\{  (X_{i}^{(1)}||X_{j}^{(0)},y_{i}^{(1)}-y_{j}^{(0)}%
):X_{j}^{(0)}\in\mathcal{C},\ X_{i}^{(1)}\in\mathcal{T}\right\}  .
\]

The motivation for constructing the above training set is that we would like
to learn a regressor
\[
y^{(1)}-y^{(0)}=g(Z_{ij},\theta),
\]
which could generalize the pairs of feature vectors in order to consider new
vectors of the form $X||X$. In other words, we would like to find the
difference $y^{(1)}-y^{(0)}$ corresponding to the concatenated vector $Z=X||X$
by using the function $g$. Here $\theta$ is a vector of the model parameters,
which depends on the chosen regression model, for example, the number of
decision trees and the number of randomly selected features in a random forest.

It should be noted that points $X_{j}^{(0)}\in\mathcal{C}$ and $X_{i}^{(1)}%
\in\mathcal{T}$ may be far from each other. This implies that the
corresponding large differences between two concatenated vectors may lead to a
regression model that is trained on vectors $Z_{ij}$ which are too far from
the required vectors $Z=X||X$. As a result, the obtained model may provide
worse results. Ideally, the best regression model should be trained by using
only pairs of identical feature vectors $Z_{ii}=X_{i}||X_{i}$. It is
impossible due to the fundamental problem of computing the causal effect.
However, for every vector $X_{i}^{(1)}\in\mathcal{T}$ from the treatment
group, we can find $k$ closest or nearest neighbor vectors $X_{j}^{(0)}%
\in\mathcal{C}$ from the control group. Let $\mathcal{C}_{i}(k)\subseteq
\mathcal{C}$ be the subset of $k$ nearest neighbors to $X_{i}^{(1)}%
\in\mathcal{T}$ from $\mathcal{C}$. In other words, for every vector
$X_{i}^{(1)}\in\mathcal{T}$, we construct a subset $\mathcal{G}_{i}%
\subset\mathcal{G}$ consisting of $k$ elements such that
\[
\mathcal{G}_{i}=\left\{  (Z_{ij},y_{i}^{(1)}-y_{j}^{(0)}):X_{j}^{(0)}%
\in\mathcal{C}_{i}(k),\ X_{i}^{(1)}\in\mathcal{T}\right\}  .
\]

The parameter $k$ can be viewed as a tuning parameter of the model. If the
number of elements in the treatment group is $n_{1}$, then the produced
training set consists of $n_{1}\cdot k$ elements. A proper choice of the
tuning parameter $k$ is very important. If we take $k=1$, then our new
training set will consists of $n_{1}$ elements which may lead to overfitting
because it is supposed that the number of patients in the treatment group is
small in comparison with the control group. If we take $k=n_{0}$, then every
element from the treatment group is concatenated with all elements from the
control group. On the one hand, we get a large training set $\mathcal{G}$
consisting of $n_{1}\cdot n_{0}$. On the other hand, the obtained training set
will contain many useless elements which are characterized by a large distance
between constituent vectors $X_{j}^{(0)}$ and $X_{i}^{(1)}$ and may
deteriorate the model.

The nearest neighbors can be found by using various distance metrics. The most
popular distances are based on the Euclidean, the Manhattan and Minkowski
($L_{p}$) metrics.%

%TCIMACRO{\FRAME{ftbpFU}{3.3408in}{2.0461in}{0pt}{\Qcb{Intuition behind the $k$
%nearest control neighbors for a treatment observation}}{\Qlb{fig:knn_cate_1}%
%}{knn_cate_1.png}{\special{ language "Scientific Word";  type "GRAPHIC";
%width 3.3408in;  height 2.0461in;  depth 0pt;  original-width 3.6149in;
%original-height 2.2035in;  cropleft "0";  croptop "1";  cropright "1";
%cropbottom "0";  filename 'KNN_CATE_1.png';file-properties "XNPEU";}}}%
%BeginExpansion
\begin{figure}
[ptb]
\begin{center}
\includegraphics[
%%natheight=2.203500in,
%%natwidth=3.614900in,
height=2.0461in,
width=3.3408in
]%
{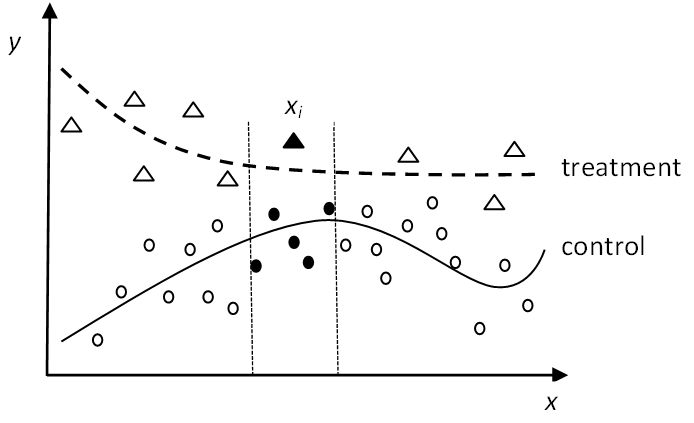}%
\caption{Intuition behind the $k$ nearest control neighbors for a treatment
observation}%
\label{fig:knn_cate_1}%
\end{center}
\end{figure}
%EndExpansion

Fig. \ref{fig:knn_cate_1} shows the observed outcomes for patients from the
control group (circles) and for patients from the treatment group (triangles).
The solid and dashed curves are the functions $g_{0}$ and $g_{1}$
corresponding to the control and treatment regressions, respectively. The
shaded triangle and 5 shaded circles illustrate a feature vector $X_{i}%
^{(1)}\in\mathcal{T}$ from the treatment group and its $5$ nearest neighbors
$X_{j}^{(0)}\in\mathcal{C}$ from the control group. The subset $\mathcal{G}%
_{i}$ in this case consists of $5$ concatenated pairs of these feature vectors.

The parameter $k$ is difficult to define especially when we consider pairs of
vectors. Moreover, for some points $X_{i}^{(1)}\in\mathcal{T}$, $k$ nearest
neighbors may include vectors $X_{j}^{(0)}\in\mathcal{C}$ which are far from
$X_{i}^{(1)}$. This may lead to incorrect results. Another way to implement
the proposed method is to restrict the largest distance between $X_{i}%
^{(1)}\in\mathcal{T}$ and $X_{j}^{(0)}\in\mathcal{C}$ by some parameter $T$,
i.e., we introduce the following constraint for constructing the set of
nearest neighbors $\mathcal{G}_{T}(i)$: $d(X_{j}^{(0)},X_{i}^{(1)})\leq T$,
i.e.,
\[
\mathcal{G}_{T}(i)=\left\{  (X_{i}^{(1)}||X_{j}^{(0)},y_{i}^{(1)}-y_{j}%
^{(0)}):d(X_{j}^{(0)},X_{i}^{(1)})\leq T\right\}  ,\ i=1,...,n_{1}.
\]
This implies that the concatenated feature vectors arise from the treatment
and control vectors which are close to each other with the upper bound $T$ of
the distance between them.

In order to take into account the different importance of concatenated
vectors, we also introduce weights $w_{ij}$ which depend on the distance
between feature vectors $X_{i}^{(1)}$ and $X_{j}^{(0)}$. Smaller values of
$d(X_{j}^{(0)},X_{i}^{(1)})$ produce larger values of weights, whereas larger
values of the distances correspond to smaller weights. We define the weights
as
\[
w_{ij}=1/d(X_{j}^{(0)},X_{i}^{(1)}).
\]

The weights of examples are used in the weighted version of the CART algorithm
to build the random forest which implements the regression constructed on the
concatenated vectors.

Suppose that the number of controls satisfying the condition $d(X_{j}%
^{(0)},X_{i}^{(1)})\leq T$ for every $i$ is $k_{i}$. Then the number of new
concatenated vectors is $k_{1}+...+k_{n_{1}}$.

In order to increase the number of concatenated feature vectors, we propose a
procedure which can be viewed as a way for augmentation of the dataset.

We assume that the number of control vectors is rather large in comparison
with the number of treatment vectors. Therefore, the first step for
implementing the augmentation procedure is to construct a regression model
$g_{0}$ by using only control feature vectors $X_{j}^{(0)}\in\mathcal{C}$,
i.e., we get $g_{0}(X^{(0)},\theta)$, where $\theta$ is a vector of
parameters. The assumption of a lot of control vectors allows us to suppose
that the regression model $g_{0}$ is precise and can be used for computing new vectors.

Then for every $X_{i}^{(1)}$ from the treatment set, we add to the initial
training set of concatenated feature vectors new vectors which are formed as
follows:%
\[
\left(  X_{l}^{(1)}||X_{l}^{(1)},y_{l}^{(1)}-g_{0}(X_{l}^{(1)},\theta)\right)
.
\]

The obtained vectors have the unit weights, i.e., $w_{ll}=1$. As a result, we
have the first augmented training set%

\[
\mathcal{G}_{f}=\left\{  X_{l}^{(1)}||X_{l}^{(1)},y_{l}^{(1)}-g_{0}%
(X_{l}^{(1)},\theta):X_{l}^{(1)}\in\mathcal{T}\right\}  .
\]

In addition to the set $\mathcal{G}_{f}$, we produce another augmented set
$\mathcal{G}_{g}(i)$ by generating $K$ random feature vectors $\Delta
X_{k}^{(1)}$ such that $d(\Delta X_{k}^{(1)},0)\leq T$, $k\in\mathcal{K}$.
Here $\mathcal{K}$ is an index set of size $K$. Now we add these vectors to
the vectors $X_{i}^{(1)}\in\mathcal{T}$ and get new vectors $\widetilde
{X}_{ik}^{(0)}=$ $X_{i}^{(1)}+\Delta X_{k}^{(1)}$. As a result, we have the
set $\mathcal{G}_{g}(i)$:
\[
\mathcal{G}_{g}(i)=\left\{  X_{i}^{(1)}||\widetilde{X}_{ik}^{(0)},y_{i}%
^{(1)}-g_{0}(\widetilde{X}_{ik}^{(0)},\theta):k\in\mathcal{K}\right\}  .
\]

The weight of every new vector from $\mathcal{G}_{g}(i)$ is $1/d(X_{i}%
^{(1)},\widetilde{X}_{ik}^{(0)})=1/d(\Delta X_{k}^{(1)},0)$. The value of $K$
is chosen such that the number of synthetic and original vectors are rather close.

Finally, we have the set of concatenated vectors
\[
\mathcal{G}=\bigcup\nolimits_{i:X_{i}^{(1)}\in\mathcal{T}}\left(
\mathcal{G}_{T}(i)\cup\mathcal{G}_{g}(i)\right)  \cup\mathcal{G}_{f},
\]
such that every element of the set $\mathcal{G}$ has its weight. The weights
of all elements are normalized. The total number of all concatenated vectors
is
\[
k_{1}+...+k_{n_{1}}+(1+K)n_{1}.
\]

By using the obtained concatenated vectors with the corresponding outcomes, we
construct the regression function $y=f\left(  X^{(1)}||X^{(0)},\vartheta
\right)  $. Here $\vartheta$ is a vector of the model parameters. Then for a
new vector $X$, the CATE\ can be determined as
\[
\tau(X)=f\left(  X||X,\vartheta\right)  .
\]

Algorithm \ref{alg:HTE_concat_1} can be viewed as a formal scheme for
computing the CATE $\tau$.

\begin{algorithm}
\caption{The algorithm for computing CATE} \label{alg:HTE_concat_1}

\begin{algorithmic}
[1]\REQUIRE Control and treatment groups $\mathcal{C}$ and $\mathcal{T}$;
parameters $K$, $T$; distance function $d$; parameters of regressions $\theta
$, $\vartheta$

\ENSURE$\tau(X)$

\STATE Construct the regression model $g_{0}(X^{(0)},\theta)$ on $\mathcal{C}$

\STATE Construct the set $\mathcal{G}_{f}=\left\{  X_{l}^{(1)}||X_{l}%
^{(1)},y_{l}^{(1)}-g_{0}(X_{l}^{(1)},\theta):X_{l}^{(1)}\in\mathcal{T}%
\right\}  $ with weights $w_{ll}=1$

\STATE For every $X_{i}^{(1)}\in\mathcal{T}$, construct the set $\mathcal{G}%
_{T}(i)=\left\{  (X_{i}^{(1)}||X_{j}^{(0)},y_{i}^{(1)}-y_{j}^{(0)}%
):d(X_{j}^{(0)},X_{i}^{(1)})\leq T\right\}  $ with weights $w_{ij}%
=1/d(X_{j}^{(0)},X_{i}^{(1)})$

\STATE Generate $K$ vectors $\Delta X_{k}^{(1)}$ under condition $d(\Delta
X_{k}^{(1)},0)\leq T$, $k\in\mathcal{K}$, and compute $\widetilde{X}%
_{ik}^{(0)}=X_{i}^{(1)}+\Delta X_{k}^{(1)}$ for every $X_{i}^{(1)}%
\in\mathcal{T}$

\STATE For every $X_{i}^{(1)}\in\mathcal{T}$, construct the set $\mathcal{G}%
_{g}(i)=\left\{  X_{i}^{(1)}||\widetilde{X}_{ik}^{(0)},y_{i}^{(1)}%
-g_{0}(\widetilde{X}_{ik}^{(0)},\theta):k\in\mathcal{K}\right\}  $ with
weights $w_{ij}=1/d(\Delta X_{k}^{(1)},0)$

\STATE$\mathcal{G}=\bigcup\nolimits_{i:X_{i}^{(1)}\in\mathcal{T}}\left(
\mathcal{G}_{T}(i)\cup\mathcal{G}_{g}(i)\right)  \cup\mathcal{G}_{f}$

\STATE Construct the regression model $y=f\left(  X^{(1)}||X^{(0)}%
,\vartheta\right)  $ on $\mathcal{G}$

\STATE$\tau(X)=f\left(  X||X,\vartheta\right)  $
\end{algorithmic}
\end{algorithm}

\section{Numerical experiments}

It is pointed out by Kunzel et al. \cite{Kunzel-etal-2018a} that the true
CATEs have to be known in order to evaluate the performance of different CATE
estimators. But the true CATEs are unknown due to the fundamental problem of
causal inference and evaluating a CATE estimator on real data is difficult
since one does not observe the true CATE. Therefore, in order to validate new
methods and to compare them with available ones, simulation experiments are
used where some known control and treatment outcome functions are taken and
random values are randomly generated in accordance with the predefined outcome functions.

In this section, we present simulation experiments evaluating the performance
of meta-models for CATE estimation. In particular, we compare the T-learner,
X-learner and Co-learner in several simulation studies by using the simulation
scheme provided by Kunzel et al. \cite{Kunzel-etal-2018a}. In the same way,
the propensity score is chosen to be constant and very small, $e(X)$ $=0.05$,
such that on average only five percents of the patients receive treatment.
Furthermore, we choose the response functions in such a way that the CATE
function is comparatively simple to estimate. For all experiments, we also use
the following identical parameters:

\begin{itemize}
\item numbers of control and treatment patients: $n_{0}=95,n_{1}=5$ and
$n_{0}=475,n_{1}=25$

\item numbers of features: $m=10$, $20$, $30$;

\item the largest distance: $T=1.5$.
\end{itemize}

The performance of the models is measured in terms of the mean squared error
(MSE). Regression functions $f\left(  X^{(1)}||X^{(0)},\vartheta\right)  $ and
$g_{0}(X^{(0)},\theta)$ are implemented by using the random forest consisting
of $1000$ decision trees. The vine-based method is used for generating random
correlation matrices \cite{Lewandowski-etal-2009}.

\subsection{Simulation experiments 1\label{subsec:simul1}}

The first simulated experiments have the following parameters:

\begin{itemize}
\item the control outcome function: $g_{0}(X)=X^{\mathrm{T}}\beta
+5I(x_{1}>0.5)$;

\item the treatment outcome function: $g_{1}(X)=X^{\mathrm{T}}\beta
+5I(x_{1}>0.5)+8\sum_{k=1}^{M}I(x_{k}>0.1)$, where $i$ is a random feature
index, $M$ is a number of feature jumps;

\item $\beta\sim Uniform\left(  [-5;5]^{m}\right)  $.
\end{itemize}

Here $I$ is the indicator function taking value $1$ if its argument is true.
It can be seen from the above that if $M=1$, then there holds $g_{1}%
(X)=g_{0}(X)+8I(x_{i}>0.1)$, and the CATE function is $\tau(X)=8I(x_{i}>0.1)$.
To evaluate the average accuracy, we perform 100 repetitions, where in each
run, we randomly select the feature index $i$ for constructing function
$g_{1}(X)$ and vector $\beta$ consisting of $m$ elements.

\begin{figure}
[ptb]
\begin{center}
\includegraphics[
%%natheight=2.650700in,
%%natwidth=8.572900in,
height=1.8611in,
width=5.9733in
]%
{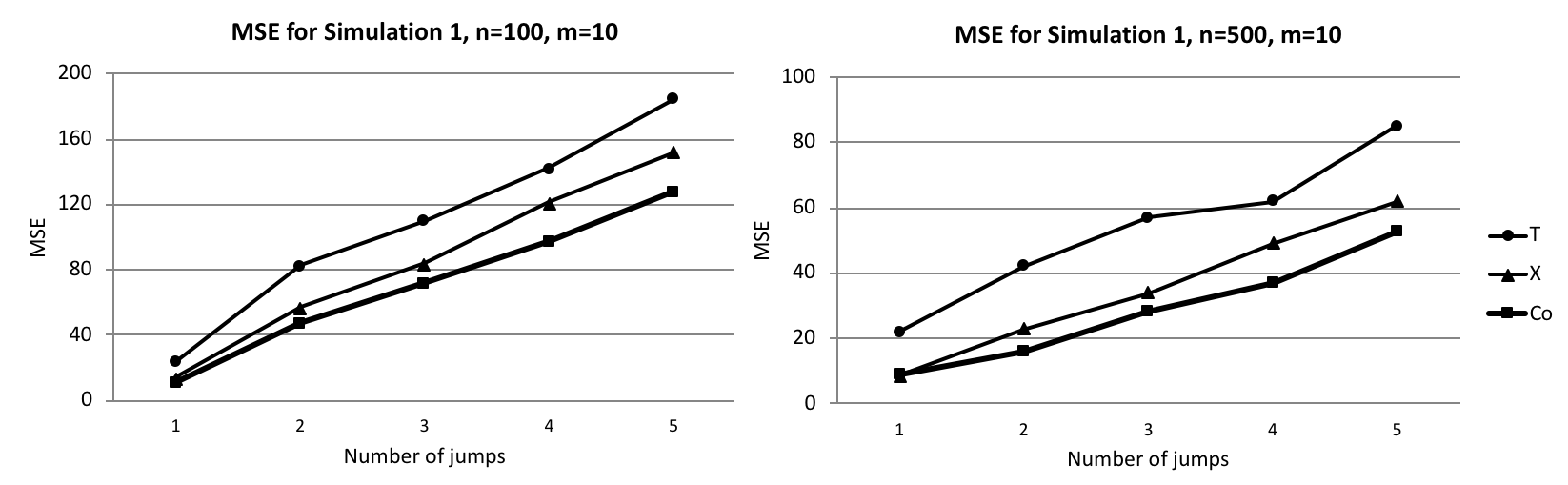}%
\caption{MSE for Simulation experiments 1 by using T-learner, X-learner,
Co-learner when $n_{0}=95,n_{1}=5$ (left) and $n_{0}=475,n_{1}=25$ (right) by
$m=10$}%
\label{fig:sim1_n100-500_m10}%
\end{center}
\end{figure}

\begin{figure}
[ptb]
\begin{center}
\includegraphics[
%%natheight=2.629900in,
%%natwidth=8.583300in,
height=1.8395in,
width=5.9586in
]%
{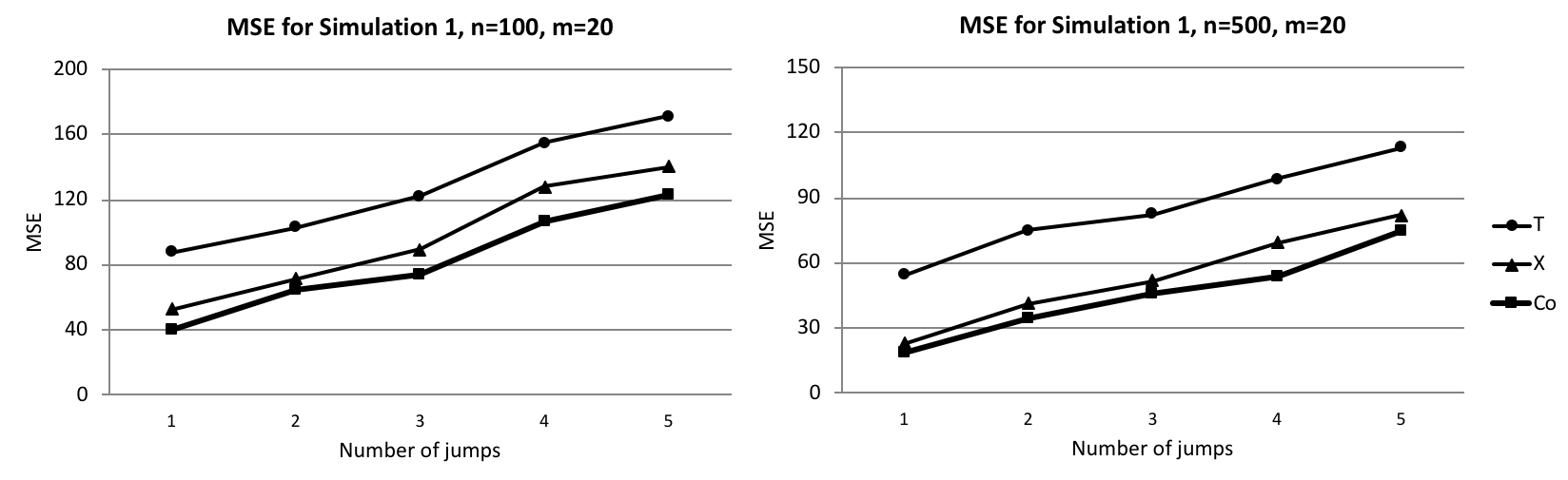}%
\caption{MSE for Simulation experiments 1 by using T-learner, X-learner,
Co-learner when $n_{0}=95,n_{1}=5$ (left) and $n_{0}=475,n_{1}=25$ (right) by
$m=20$}%
\label{fig:sim1_n100-500_m20}%
\end{center}
\end{figure}

\begin{figure}
[ptb]
\begin{center}
\includegraphics[
%%natheight=2.640300in,
%%natwidth=8.562500in,
height=1.8325in,
width=5.9015in
]%
{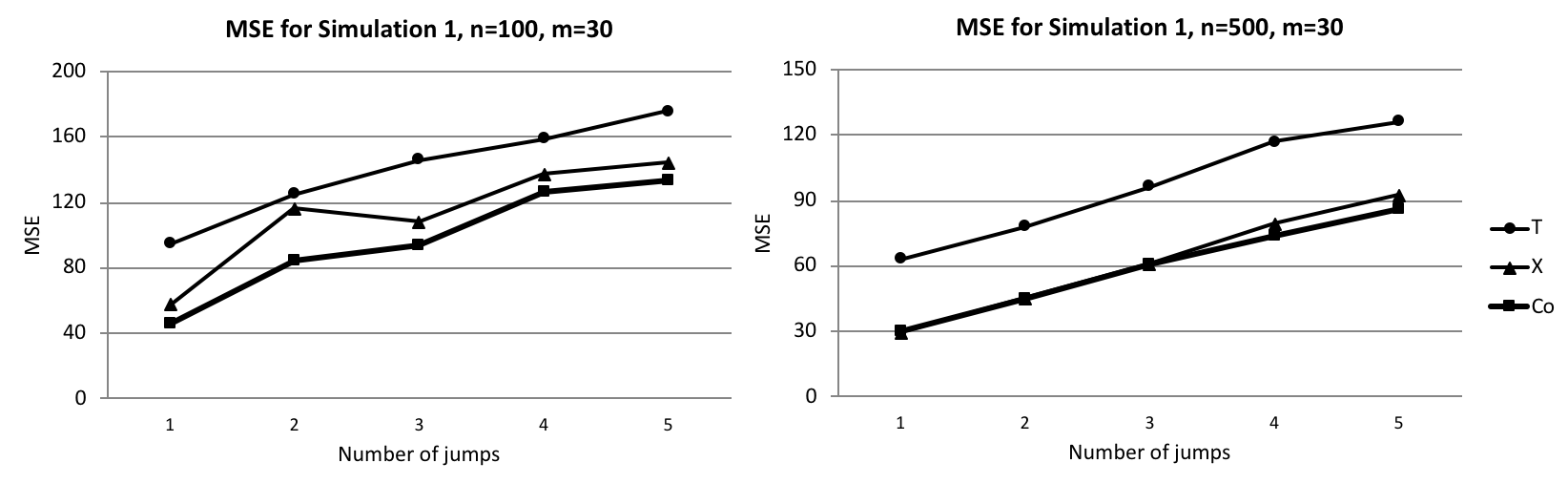}%
\caption{MSE for Simulation experiments 1 by using T-learner, X-learner,
Co-learner when $n_{0}=95,n_{1}=5$ (left) and $n_{0}=475,n_{1}=25$ (right) by
$m=30$}%
\label{fig:sim1_n100-500_m30}%
\end{center}
\end{figure}

In order to compare three models: T-learner, X-learner, Co-learner, we perform
several experiments with different parameters, in particular, we consider how
values of the MSE depend on the number of generated jumps of the CATE
function. The corresponding results are shown in Figs.
\ref{fig:sim1_n100-500_m10}-\ref{fig:sim1_n100-500_m30}. The curves with
circle, triangular, and squared markers correspond to the T-learner,
X-learner, and Co-learner, respectively. Moreover, the curves of the MSE for
the Co-learner are depicted to be thick in order to distinguish them from
curves corresponding to other models.

It can be seen from the figures that the Co-learner outperforms T-learner and
X-learner. We have to point out that the difference between values of the MSE
for the X-learner and the Co-learner decreases with the total number of
patients. This implies that the proposed model provides better results in
comparison with other models when corresponding datasets are rather small. In
most experiments, the difference between values of the MSE for the X-learner
and the Co-learner increases with the number of the CATE function jumps. This
is an important property which shows that the proposed meta-model outperforms
other considered models when the treatment outcome function becomes to be complicated.%

%TCIMACRO{\FRAME{ftbpFU}{5.9733in}{1.8611in}{0pt}{\Qcb{MSE for Simulation
%experiments 1 by using T-learner, X-learner, Co-learner when $n_{0}%
%=95,n_{1}=5$ (left) and $n_{0}=475,n_{1}=25$ (right) by $m=10$}}%
%{\Qlb{fig:sim1_n100-500_m10}}{sim1_n100-500_m10.png}%
%width 5.9733in;  height 1.8611in;  depth 0pt;  original-width 8.5729in;
%original-height 2.6507in;  cropleft "0";  croptop "1";  cropright "1";
%cropbottom "0";  filename 'Sim1_n100-500_m10.png';file-properties "XNPEU";}}}%
%BeginExpansion
%EndExpansion
%

%TCIMACRO{\FRAME{ftbpFU}{5.9586in}{1.8395in}{0pt}{\Qcb{MSE for Simulation
%experiments 1 by using T-learner, X-learner, Co-learner when $n_{0}%
%=95,n_{1}=5$ (left) and $n_{0}=475,n_{1}=25$ (right) by $m=20$}}%
%{\Qlb{fig:sim1_n100-500_m20}}{sim1_n100-500_m20.png}%
%width 5.9586in;  height 1.8395in;  depth 0pt;  original-width 8.5833in;
%original-height 2.6299in;  cropleft "0";  croptop "1";  cropright "1";
%cropbottom "0";  filename 'Sim1_n100-500_m20.png';file-properties "XNPEU";}}}%
%BeginExpansion
%EndExpansion
%

%TCIMACRO{\FRAME{ftbpFU}{5.9015in}{1.8325in}{0pt}{\Qcb{MSE for Simulation
%experiments 1 by using T-learner, X-learner, Co-learner when $n_{0}%
%=95,n_{1}=5$ (left) and $n_{0}=475,n_{1}=25$ (right) by $m=30$}}%
%{\Qlb{fig:sim1_n100-500_m30}}{sim1_n100-500_m30.png}%
%width 5.9015in;  height 1.8325in;  depth 0pt;  original-width 8.5625in;
%original-height 2.6403in;  cropleft "0";  croptop "1";  cropright "1";
%cropbottom "0";  filename 'Sim1_n100-500_m30.png';file-properties "XNPEU";}}}%
%BeginExpansion
%EndExpansion

\begin{figure}
[ptb]
\begin{center}
\includegraphics[
%%natheight=2.619500in,
%%natwidth=8.572900in,
height=1.8256in,
width=5.93in
]%
{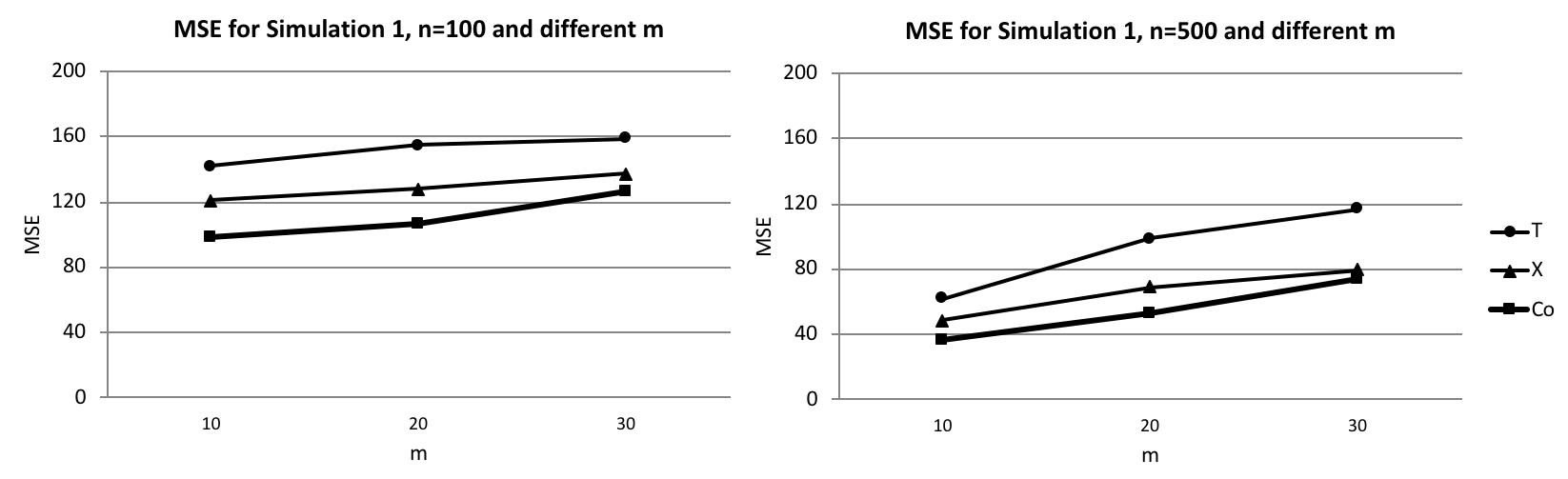}%
\caption{MSE for Simulation experiments 1 by using T-learner, X-learner,
Co-learner when $n_{0}=95,n_{1}=5$ (left) and $n_{0}=475,n_{1}=25$ (right) by
different $m$ and $4$ jumps of the CATE function}%
\label{fig:sim1_n100-500_diff_m}%
\end{center}
\end{figure}

Fig. \ref{fig:sim1_n100-500_diff_m} shows how values of the MSE explicitly
depend on the number of features $m$ by $n_{0}=95,n_{1}=5$ (left curves) and
$n_{0}=475,n_{1}=25$ (right curves). We consider the case when the number of
jumps of the CATE function is $4$. It can be concluded from the experiments
that there is no a significant difference between behavior of meta-models,
i.e., values of the MSE for all meta-models increase with the feature number.%

%TCIMACRO{\FRAME{ftbpFU}{5.93in}{1.8256in}{0pt}{\Qcb{MSE for Simulation
%experiments 1 by using T-learner, X-learner, Co-learner when $n_{0}%
%=95,n_{1}=5$ (left) and $n_{0}=475,n_{1}=25$ (right) by different $m$ and $4$
%jumps of the CATE function}}{\Qlb{fig:sim1_n100-500_diff_m}}%
%{sim1_n100-500_diff_m.png}{\special{ language "Scientific Word";
%type "GRAPHIC";  maintain-aspect-ratio TRUE;  display "USEDEF";
%valid_file "F";  width 5.93in;  height 1.8256in;  depth 0pt;
%original-width 8.5729in;  original-height 2.6195in;  cropleft "0";
%filename 'Sim1_n100-500_diff_m.png';file-properties "XNPEU";}}}%
%BeginExpansion
%EndExpansion

\begin{figure}
[ptb]
\begin{center}
\includegraphics[
%%natheight=2.629900in,
%%natwidth=8.521000in,
height=1.8533in,
width=5.9646in
]%
{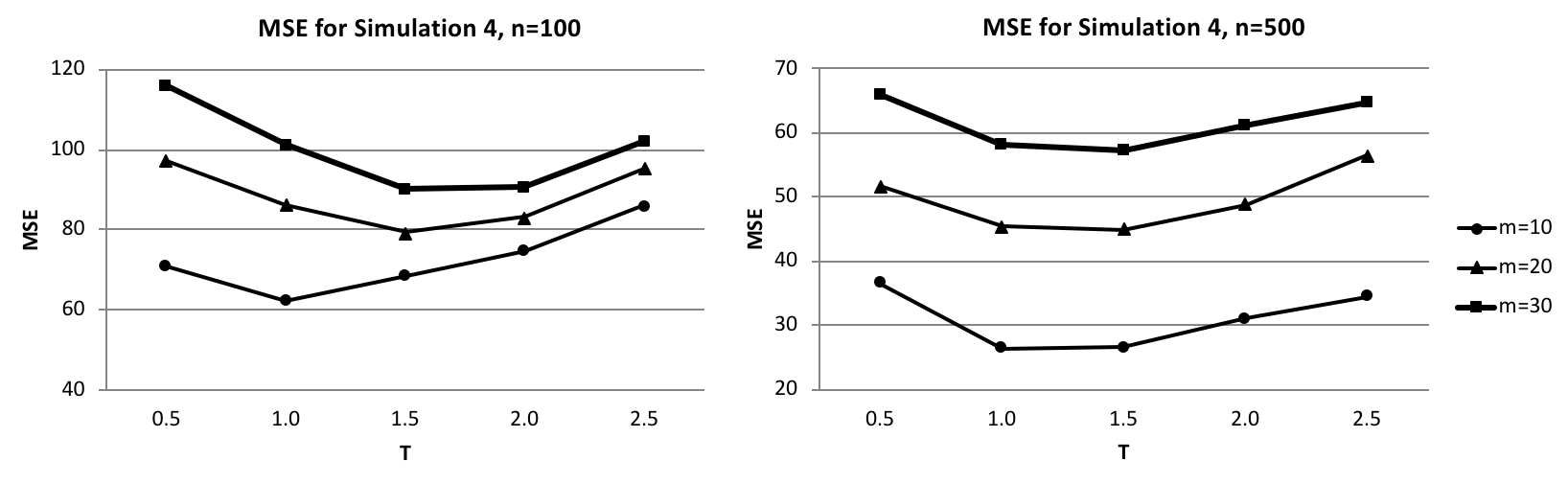}%
\caption{MSE for Co-learner in Simulation experiments 1 as a function of the
largest distance $T$ when $n_{0}=95,n_{1}=5$ (left) and $n_{0}=475,n_{1}=25$
(right) by different $m$}%
\label{fig:sim4}%
\end{center}
\end{figure}

In order to study how the upper bound $T$ for the distance between
concatenated treatment and control vectors (constructing sets $\mathcal{G}%
_{T}(i)$) impacts on the proposed model MSE, we perform the corresponding
experiments with parameters of Simulation 1 and different $T$ for
$m=10,20,30$. Three jumps of the CATE function are simulated. The
corresponding results are shown in Fig. \ref{fig:sim4}. It is interesting to
note from Fig. \ref{fig:sim4} that, for every $m$, there is an optimal value
of $T$ which leads to the smallest MSE. In particular, the smallest values of
the MSE for $m=10,20,30$ and by $n_{0}=95,n_{1}=5$ (left curves) are achieved
by $T=1$, $1.5$ and $2$, respectively. If we consider the case $n_{0}%
=475,n_{1}=25$ (right curves), then the optimal values of $T$ are $1$, $1.5$
and $1.5$.%

%TCIMACRO{\FRAME{ftbpFU}{5.9646in}{1.8533in}{0pt}{\Qcb{MSE for Co-learner in
%Simulation experiments 1 as a function of the largest distance $T$ when
%$n_{0}=95,n_{1}=5$ (left) and $n_{0}=475,n_{1}=25$ (right) by different $m$}%
%}{\Qlb{fig:sim4}}{sim4.png}{\special{ language "Scientific Word";
%type "GRAPHIC";  maintain-aspect-ratio TRUE;  display "USEDEF";
%valid_file "F";  width 5.9646in;  height 1.8533in;  depth 0pt;
%original-width 8.521in;  original-height 2.6299in;  cropleft "0";
%filename 'Sim4.png';file-properties "XNPEU";}}}%
%BeginExpansion
%EndExpansion

\begin{figure}
[ptb]
\begin{center}
\includegraphics[
%%natheight=2.640300in,
%%natwidth=8.447500in,
height=1.8758in,
width=5.9612in
]%
{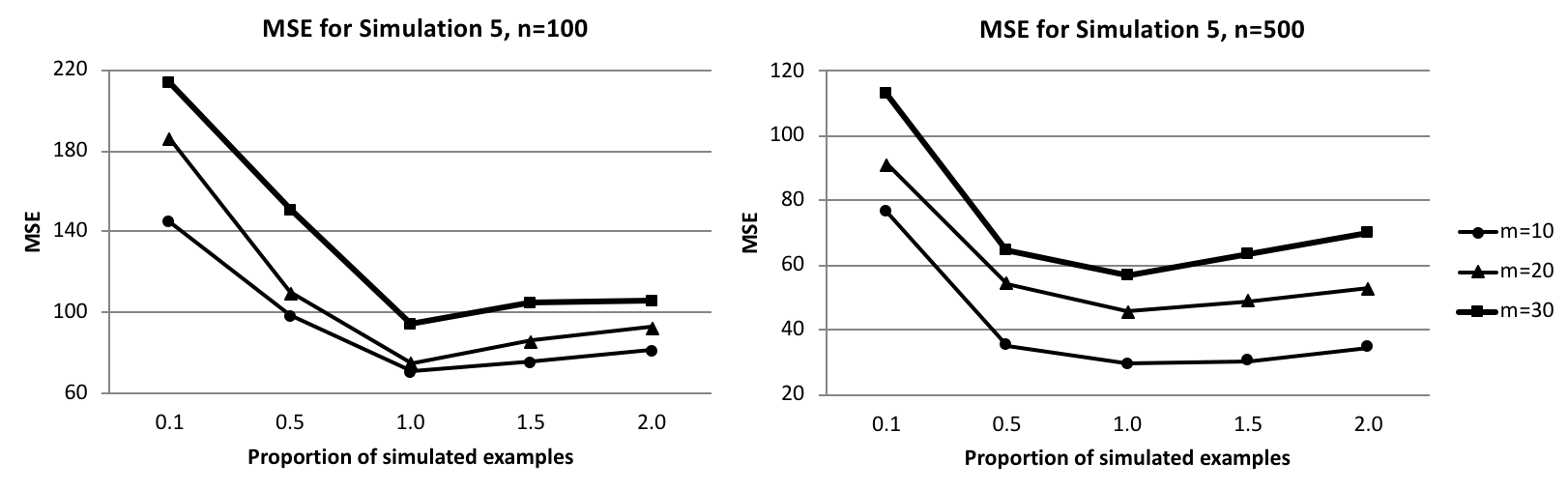}%
\caption{MSE for Co-learner in Simulation experiments 1 as a function of the
proportion of simulated examples when $n_{0}=95,n_{1}=5$ (left) and
$n_{0}=475,n_{1}=25$ (right) by different $m$}%
\label{fig:sim5}%
\end{center}
\end{figure}

By using parameters from the previous experiment, we study how the proportion
of simulated examples (constructing sets $\mathcal{G}_{g}(i)$) impacts on the
proposed model MSE. The proportion is defined as the ratio of $K\cdot n_{1}$
generated random feature vectors and the number of vectors from the set
$\mathcal{G}_{T}(1)\cup...\cup\mathcal{G}_{T}(n_{1})$. The results are shown
in Fig. \ref{fig:sim5}. We again see that there are optimal values of the
proportion providing the smallest values of MSE. In contrast to the above
experiment, we have the optimal value $1$ for all cases.%

%TCIMACRO{\FRAME{ftbpFU}{5.9612in}{1.8758in}{0pt}{\Qcb{MSE for Co-learner in
%Simulation experiments 1 as a function of the proportion of simulated examples
%when $n_{0}=95,n_{1}=5$ (left) and $n_{0}=475,n_{1}=25$ (right) by different
%$m$}}{\Qlb{fig:sim5}}{sim5.png}{\special{ language "Scientific Word";
%type "GRAPHIC";  maintain-aspect-ratio TRUE;  display "USEDEF";
%valid_file "F";  width 5.9612in;  height 1.8758in;  depth 0pt;
%original-width 8.4475in;  original-height 2.6403in;  cropleft "0";
%filename 'Sim5.png';file-properties "XNPEU";}}}%
%BeginExpansion
%EndExpansion

\subsection{Simulation experiments 2}

The second simulated experiments use non-linear outcome functions, but the
CATE function is the same as in Simulation 2. The experiments have the
following parameters:

\begin{itemize}
\item the control outcome function: $g_{0}(X)=\varepsilon(x_{1})\varepsilon
(x_{2})/2$; where $\varepsilon(x)=2\left(  1+\exp\left(  -12(x-0.5)\right)
\right)  ^{-1}$

\item the treatment outcome function: $g_{1}(X)=\varepsilon(x_{1}%
)\varepsilon(x_{2})/2+8I(x_{i}>0.1)$, where $i$ is a random feature index.
\end{itemize}

It can be seen from the above that we have again $g_{1}(X)=g_{0}%
(X)+8\sum_{k=1}^{M}I(x_{k}>0.1)$, and the CATE function by $M=1$ is again
$\tau(X)=8I(x_{i}>0.1)$.

Since the range of the MSE values for different numbers of the CATE function
jumps significantly exceeds the difference between the MSE values of the
meta-learners, then we illustrate obtained numerical results by means of
Tables \ref{t:HTE_Concat_2_1}-\ref{t:HTE_Concat_2_3}. Every table is defined
by the number of features $m$. We again study how values of the MSE depend on
the number of generated jumps of the CATE function by different parameters
$n_{0}$ and $n_{1}$. It can be seen from Tables \ref{t:HTE_Concat_2_1}%
-\ref{t:HTE_Concat_2_3} that the Co-learner outperforms the T-learner and the
X-learner in all presented cases. It is also interesting to note that the
T-learner outperforms the X-learner by some parameters. However, these
learners do not perform the Co-learner in all cases.%

%TCIMACRO{\TeXButton{B}{\begin{table}[tbp] \centering}}%
%BeginExpansion
\begin{table}[tbp] \centering
%EndExpansion
\caption{The MSE for Simulation  experiments 2 by $m=10$}%
\begin{tabular}
[c]{ccccccccccc}\hline
& \multicolumn{5}{c}{$n_{0}=95,n_{1}=5$} & \multicolumn{5}{c}{$n_{0}%
=475,n_{1}=25$}\\\hline
Jumps & $1$ & $2$ & $3$ & $4$ & $5$ & $1$ & $2$ & $3$ & $4$ & $5$\\\hline
T & $16.9$ & $36.8$ & $62.3$ & $92.1$ & $126$ & $2.10$ & $11.7$ & $20.9$ &
$34.8$ & $46.7$\\\hline
X & $16.6$ & $36.6$ & $62.0$ & $91.4$ & $126$ & $1.82$ & $11.2$ & $20.4$ &
$34.4$ & $46.4$\\\hline
Co & $15.1$ & $34.1$ & $57.5$ & $87.6$ & $119$ & $1.49$ & $9.89$ & $20.0$ &
$33.8$ & $44.5$\\\hline
\end{tabular}
\label{t:HTE_Concat_2_1}%
%TCIMACRO{\TeXButton{E}{\end{table}}}%
%BeginExpansion
\end{table}%
%EndExpansion
%

%TCIMACRO{\TeXButton{B}{\begin{table}[tbp] \centering}}%
%BeginExpansion
\begin{table}[tbp] \centering
%EndExpansion
\caption{The MSE for Simulation  experiments 2 by $m=20$}%
\begin{tabular}
[c]{ccccccccccc}\hline
& \multicolumn{5}{c}{$n_{0}=95,n_{1}=5$} & \multicolumn{5}{c}{$n_{0}%
=475,n_{1}=25$}\\\hline
Jumps & $1$ & $2$ & $3$ & $4$ & $5$ & $1$ & $2$ & $3$ & $4$ & $5$\\\hline
T & $18.2$ & $32.9$ & $59.6$ & $72.5$ & $108$ & $1.60$ & $13.4$ & $26.3$ &
$40.4$ & $57.8$\\\hline
X & $18.2$ & $33.1$ & $59.9$ & $72.6$ & $110$ & $1.33$ & $9.45$ & $26.1$ &
$40.1$ & $58.3$\\\hline
Co & $1.70$ & $31.0$ & $56.7$ & $69.3$ & $103$ & $0.91$ & $5.30$ & $25.3$ &
$39.5$ & $56.4$\\\hline
\end{tabular}
\label{t:HTE_Concat_2_2}%
%TCIMACRO{\TeXButton{E}{\end{table}}}%
%BeginExpansion
\end{table}%
%EndExpansion
%

%TCIMACRO{\TeXButton{B}{\begin{table}[tbp] \centering}}%
%BeginExpansion
\begin{table}[tbp] \centering
%EndExpansion
\caption{The MSE for Simulation  experiments 2 by $m=30$}%
\begin{tabular}
[c]{ccccccccccc}\hline
& \multicolumn{5}{c}{$n_{0}=95,n_{1}=5$} & \multicolumn{5}{c}{$n_{0}%
=475,n_{1}=25$}\\\hline
Jumps & $1$ & $2$ & $3$ & $4$ & $5$ & $1$ & $2$ & $3$ & $4$ & $5$\\\hline
T & $18.2$ & $41.8$ & $62.1$ & $87.4$ & $104$ & $1.94$ & $14.6$ & $29.4$ &
$47.9$ & $64.1$\\\hline
X & $18.3$ & $42.0$ & $63.0$ & $87.6$ & $105$ & $1.67$ & $13.3$ & $29.6$ &
$50.4$ & $64.8$\\\hline
Co & $17.3$ & $41.9$ & $60.5$ & $87.1$ & $101$ & $1.08$ & $12.5$ & $28.6$ &
$46.7$ & $62.9$\\\hline
\end{tabular}
\label{t:HTE_Concat_2_3}%
%TCIMACRO{\TeXButton{E}{\end{table}}}%
%BeginExpansion
\end{table}%
%EndExpansion

\subsection{Simulation experiments 3\label{subsec:simul3}}

The third simulated experiments have the following parameters:

\begin{itemize}
\item the control outcome function: $g_{0}(X)=\varepsilon(x_{1})\varepsilon
(x_{2})/2$; where $\varepsilon(x)=2\left(  1+\exp\left(  -12(x-0.5)\right)
\right)  ^{-1}$

\item the treatment outcome function: $g_{1}(X)=-\varepsilon(x_{1}%
)\varepsilon(x_{2})/2$.
\end{itemize}

It can be seen from the above that the $g_{1}(X)=-g_{0}(X)$, and the CATE
function is $\tau(X)=-\varepsilon(x_{1})\varepsilon(x_{2})$.

In this simulation experiment, we have quite different results which are shown
in Table \ref{t:HTE_Concat_3_1}. The Co-learner provides worse results in
comparison with other models. It is interesting to point out that the
Co-learner illustrated outperforming results when the CATE\ function had jumps
which correspond to a strong reaction of patients with specific sets of
features to treatment. However, when the treatment \textquotedblleft
smoothly\textquotedblright\ impacts on all patients from the treatment group,
then the X-learner and even T-learner may be preferable. The jumping outcome
function means that the treatment start to impact on patients with certain
values of some features. For example, some medicine comes into effect when the
patient age is smaller than 50.%

%TCIMACRO{\TeXButton{B}{\begin{table}[tbp] \centering}}%
%BeginExpansion
\begin{table}[tbp] \centering
%EndExpansion
\caption{The MSE for Simulation experiments 3}%
\begin{tabular}
[c]{ccccccc}\hline
& \multicolumn{3}{c}{$n_{0}=95,n_{1}=5$} & \multicolumn{3}{c}{$n_{0}%
=475,n_{1}=25$}\\\hline
$m$ & $10$ & $20$ & $30$ & $10$ & $20$ & $30$\\\hline
T & $0.856$ & $0.974$ & $1.07$ & $0.580$ & $0.613$ & $0.653$\\\hline
X & $0.601$ & $0.717$ & $0.880$ & $0.283$ & $0.305$ & $0.321$\\\hline
Co & $1.15$ & $1.24$ & $1.32$ & $0.993$ & $1.02$ & $1.10$\\\hline
\end{tabular}
\label{t:HTE_Concat_3_1}%
%TCIMACRO{\TeXButton{E}{\end{table}}}%
%BeginExpansion
\end{table}%
%EndExpansion

\section{Co-learner and the feature bagging}

Let us consider a simple modification of the Co-learner, which is based on
applying the random subspace method \cite{Ho-1998} or feature bagging. The
feature bagging is an ensemble-based method which chooses some subset of
features for constructing every individual model in the ensemble.

Let $K_{0}=\{k_{1},...,k_{l}\}$ and $K_{1}=\{k_{l+1},...,k_{l+t}\}$ be two
ordered index sets of $l$ and $t$ selected features such that $1\leq k_{j}\leq
m$, $j=1,...,l+t$, and $k_{j}\leq k_{j+1}$. Here $l<m$ and $t<m$ are tuning
parameters. We denote vectors $X_{j,K_{0}}^{(0)}\subseteq X_{j}^{(0)}$ and
$X_{i,K_{1}}^{(1)}\subseteq X_{i}^{(1)}$ as parts of vectors $X_{j}^{(0)}%
\in\mathcal{C}$ and $X_{i}^{(1)}\in\mathcal{T}$ in accordance with index sets
$K_{0}$ and $K_{1}$, respectively.

For all vectors $X_{j,K_{0}}^{(0)}$ and $X_{i,K_{1}}^{(1)}$, we find
concatenated vectors $X_{i,K_{1}}^{(1)}||X_{j,K_{0}}^{(0)}$. Denote also
$K=\{K_{1},k_{1}+m,...,k_{l}+m\}$. Then we can construct the regression model
$g_{0}(X_{K}^{(0)},\theta)$ and sets $\mathcal{G}_{f,K}$, $\mathcal{G}%
_{T,K}(i)$, $\mathcal{G}_{g,K}(i)$, which differ from the similar model
$g_{0}(X^{(0)},\theta)$ and the sets $\mathcal{G}_{f}$, $\mathcal{G}_{T}(i)$,
$\mathcal{G}_{g}(i)$ (see Algorithm \ref{alg:HTE_concat_1}) by the vectors
$X_{j,K_{0}}^{(0)}$ and $X_{i,K_{1}}^{(1)}$ in place of vectors $X_{j}^{(0)}$
and $X_{i}^{(1)}$. Finally, we compute the CATE $\tau_{K}(X)$ by using
Algorithm \ref{alg:HTE_concat_1} for the index sets $K_{0}$ and $K_{1}$.

By randomly selecting $N$ times the index sets $K_{0}$ and $K_{1}$ and by
using Algorithm \ref{alg:HTE_concat_1}, we get a set of the corresponding CATE
values $\tau_{K(i)}(X)$, $i=1,...,N$. Here $K(i)$ is the resulting index set
obtained by the $i$-th generating the random sets $K_{0}$ and $K_{1}$. The
final CATE value can be found by averaging the obtained CATE values $\tau
_{K}(X)$, i.e., there holds
\[
\tau(X)=N^{-1}\sum_{i=1}^{N}\tau_{K(i)}(X).
\]

This simple and obvious modification of the Co-learner may lead to better results.

\subsection{Simulation experiments 1}

First, we perform numerical experiments in accordance with Simulation 3 (see
Subs.\ref{subsec:simul3}), i.e., the control outcome function is
$g_{0}(X)=\varepsilon(x_{1})\varepsilon(x_{2})/2$, where $\varepsilon
(x)=2\left(  1+\exp\left(  -12(x-0.5)\right)  \right)  ^{-1}$, and the
treatment outcome function is $g_{1}(X)=-\varepsilon(x_{1})\varepsilon
(x_{2})/2$. Numbers of selected features are taken identical $l=t=2m/3$, i.e.,
two thirds of all features are used for training the regression models.

Results of numerical experiments are shown in Table \ref{t:HTE_Concat_4},
where the MSE measures are shown for the T-learner, X-learner, and Co-learner
under two conditions: numbers of control and treatment patients are
$n_{0}=95,n_{1}=5$ and $n_{0}=475,n_{1}=25$. Moreover, the corresponding
measures are shown for the Co-learner with the feature bagging, which is
denoted in Table \ref{t:HTE_Concat_4} as Co-B. Results in Table
\ref{t:HTE_Concat_4} for the Co-learner with the feature bagging are provided
by two numbers of the bagging iterations $N=10$ and $N=100$.

It is interesting to see from the results that the Co-learner with the feature
bagging outperforms the original Co-learner by $N=10$. However, it shows worse
results by $N=100$. Moreover, we again observe that Co-learners with the
feature bagging as well as the original Co-learner is inferior to the
X-learner and even T-learner. In other words, the feature bagging could not
significantly improve the proposed model.%

%TCIMACRO{\TeXButton{B}{\begin{table}[tbp] \centering}}%
%BeginExpansion
\begin{table}[tbp] \centering
%EndExpansion
\caption{The MSE for Simulation experiments 1 with feature bagging by $N=10$ and $N=100$}%
\begin{tabular}
[c]{ccccccc}\hline
& \multicolumn{3}{c}{$n_{0}=95,n_{1}=5$} & \multicolumn{3}{c}{$n_{0}%
=475,n_{1}=25$}\\\hline
$m$ & $10$ & $20$ & $30$ & $10$ & $20$ & $30$\\\hline
T & $0.856$ & $0.974$ & $1.07$ & $0.580$ & $0.613$ & $0.653$\\\hline
X & $0.601$ & $0.717$ & $0.880$ & $0.283$ & $0.305$ & $0.321$\\\hline
Co & $1.15$ & $1.24$ & $1.32$ & $0.993$ & $1.02$ & $1.10$\\\hline
Co-B by $N=10$ & $1.07$ & $1.12$ & $1.17$ & $0.816$ & $0.882$ & $0.91$\\\hline
Co-B by $N=100$ & $1.06$ & $1.14$ & $1.17$ & $1.17$ & $1.04$ & $0.95$\\\hline
\end{tabular}
\label{t:HTE_Concat_4}%
%TCIMACRO{\TeXButton{E}{\end{table}}}%
%BeginExpansion
\end{table}%
%EndExpansion

\subsection{Simulation experiments 2}

Quite different results are obtained when we perform numerical experiments
which are similar to Simulation experiments 1 (see Subs. \ref{subsec:simul1}).
However, in contrast to the previous simulation experiments, we complicate the
treatment output function. In particular, the control outcome function is the
same, it is $g_{0}(X)=X^{\mathrm{T}}\beta+5I(x_{1}>0.5)$, but the treatment
outcome function is $g_{1}(X)=X^{\mathrm{T}}\beta+5I(x_{1}>0.5)+8\sum
_{k=1}^{M}I(x_{k}>0.1)$ now, where $M=3$. It can be seen from the treatment
output function that three randomly selected features are changed in experiments.%

%TCIMACRO{\TeXButton{B}{\begin{table}[tbp] \centering}}%
%BeginExpansion
\begin{table}[tbp] \centering
%EndExpansion
\caption{The MSE for Simulation experiments 2 with feature bagging by $N=10$ and $N=100$}%
\begin{tabular}
[c]{ccccccc}\hline
& \multicolumn{3}{c}{$n_{0}=95,n_{1}=5$} & \multicolumn{3}{c}{$n_{0}%
=475,n_{1}=25$}\\\hline
$m$ & $10$ & $20$ & $30$ & $10$ & $20$ & $30$\\\hline
T & $110$ & $122$ & $146$ & $56.9$ & $82.6$ & $96.4$\\\hline
X & $83.7$ & $89.3$ & $108$ & $34.4$ & $51.6$ & $61.0$\\\hline
Co & $71.6$ & $73.6$ & $94.2$ & $28.3$ & $46.1$ & $60.7$\\\hline
Co-B by $N=10$ & $74.5$ & $73.8$ & $82.4$ & $30.2$ & $45.4$ & $54.7$\\\hline
Co-B by $N=100$ & $74.3$ & $73.1$ & $77.8$ & $30.0$ & $44.6$ & $51.9$\\\hline
\end{tabular}
\label{t:HTE_Concat_5}%
%TCIMACRO{\TeXButton{E}{\end{table}}}%
%BeginExpansion
\end{table}%
%EndExpansion

Results of experiments are given in Table \ref{t:HTE_Concat_5}. They are very
interesting. First of all, one can see that the Co-learner with the feature
bagging outperforms the X-learner as well as the T-learner. Moreover, it is
outperforms the original Co-learner for most parameters. At the same time, we
can see that the largest difference between MSE measures are observed by large
numbers of features. Moreover, the Co-learner with the feature bagging is
improved with increase of $N$. Hence, we can conclude that the Co-learner with
the feature bagging should be used when examples in a dataset are
characterized by a large number of features. For example, if the difference
between MSE measures by $m=30$ and by $m=20$ ($n_{0}=95$, $n_{1}=5$) for the
X-learner is $\allowbreak18.7$, then the same differences for the Co-learner
with the feature bagging by $N=10$ and $N=100$ are $8.6$ and $4.7$. These
outperforming results illustrate the efficiency of the proposed meta-learner
with feature bagging.\qquad

\section{Conclusion}

An approach to solving the CATE problem has been proposed. The main idea is to
construct a new dataset from the treatment and control groups by means of
concatenating the original vectors. Moreover, the obtained dataset is extended
by generating new feature vectors in accordance with predefined rules. By
developing the model, we aimed to avoid constructing the regression function
$g_{1}(X)$ for the treatment group in order to take into account the fact that
the number of patients from the treatment group may be very small.

The proposed model is simple from the computation point of view and its
complexity does not differ from the T-learner and the X-learner. Numerical
simulation experiments have illustrated the outperformance of the Co-learner
for some datasets. At the same time, some numerical experiments have shown
that the Co-learner may provide worse results in comparison with the T-learner
and the X-learner (see, for example, Simulation 3). This fact states a task of
modifying the proposed model in order to overcome this shortcoming in future.

It should be noted that we used a certain set of learning parameters. For
example, the Euclidean distance was used in order to compute weights and to
determine the number of nearest neighbors from the control group. However, we
can study various ways for defining the distance, including kernels, and other
parameters. This is a direction for further research. Another problem is that
the assumption of a large control group may be also violated in some
applications. This implies that robust models taking into account the lack of
sufficient information about controls as well as treatments should be
developed. One of the ideas for solving the problem is to apply the imprecise
probability theory \cite{Walley91} and its imprecise statistical models.
However, this is also a direction for further research.

\section*{Acknowledgement}

This work is supported by the Russian Science Foundation under grant 18-11-00078.

%\bibliographystyle{plain}
%\bibliography{Autoencoder,Boosting,Classif_bib,Deep_Forest,IntervalClass,MYBIB,MYUSE,Survival_analysis,Transf_Learn}

\end{document}